# Moving Beyond Next-Token Prediction: Transformers are Context-Sensitive Language Generators

Phill Kyu Rhee

*Inha University 100 Inha Ro, Incheon, 2222 South Korea*

*pkrhee@inha.ac.kr*

**(Abstract)** Large Language Models (LLMs), powered by Transformers, have demonstrated human-like intelligence capabilities, yet their underlying mechanisms remain poorly understood. This paper presents a novel framework for interpreting LLMs as probabilistic left context-sensitive languages (CSLs) generators. We hypothesize that Transformers can be effectively decomposed into three fundamental components: context windows, attention mechanisms, and autoregressive generation frameworks. This decomposition allows for the development of more flexible and interpretable computational models, moving beyond the traditional view of attention and autoregression as inseparable processes. We argue that next-token predictions can be understood as probabilistic, dynamic approximations of left CSL production rules, providing an intuitive explanation for how simple token predictions can yield human-like intelligence outputs. Given that all CSLs are left context-sensitive (Penttonen, 1974), we conclude that Transformers stochastically approximate CSLs, which are widely recognized as models of human-like intelligence. This interpretation bridges the gap between Formal Language Theory and the observed generative power of Transformers, laying a foundation for future advancements in generative AI theory and applications. Our novel perspective on Transformer architectures will foster a deeper understanding of LLMs and their future potentials.

## 1. INTRODUCTION

Transformers, the deep learning architecture powering state-of-the-art generative AI models like GPT (OpenAI, 2024), Llama (Meta, 2025), and Gemini (Google, 2025), have not only revolutionized text generation but also but also a spectrum of domains, including multimodal content creation (audio, image, video), cellular analysis, protein structure prediction, and medical diagnostics. Recent advancements have extended their applications to robotics, enhancing perception, planning, and control in autonomous systems, further illustrates their transformative potential (Sanghai et al., 2024). Despite architectural variations, Transformers fundamentally operate on a simple principle: autoregressive next-word prediction (Vaswani et al., 2017). Starting with a text prompt, they iteratively predict the most probable next word, constructing coherent sequences.

The core innovation of Transformers lies in their self-attention mechanisms, which effectively capture long-range dependencies and have sparked extensive research into their remarkable capabilities (Devlin et al., 2019). Furthermore, their scalability has enabled breakthroughs in domains ranging from computer vision to computational biology (Yin et al., 2024). Nevertheless, the precise mechanisms enabling Transformers to exhibit human-like intelligence remain elusive, leaving room for further exploration. Assertions that LLMs are merely "stochastic parrots" (Bender et al., 2021) fail to account for the complexity of their underlying mechanisms and emergent capabilities (Grzankowski et al., 2024).

While Transformers have been theoretically shown to be Turing-complete (Perez et al., 2019), their practical limitations, particularly regarding scalability and efficiency, are not well-documented. Some research indicates that Transformers perform inferiorly compared to competitive models like RNNs and LSTMs on tasks involving the recognition of formal languages (Hahn, 2020). Furthermore, recent research suggests that self-attention mechanisms in Transformers may have theoretical limitations in learning certain regular and context-free languages (Bhattamishra et al., 2020a). However, such critiques do not diminish Transformers' practical effectiveness, as evidenced by their remarkable successes in practical natural language processing. This discrepancy suggests potential deep differences in the interpretation of Transformers' capabilities from the perspective of formal language theory, raising significant questions about their architectural interpretation and motivating the exploration of alternative theoretical or experimental foundations. These mismatches have spurred considerable research investigating the mechanisms of Transformers to bridge the gap between formal language theories and the practical and commercial successes of LLMs.

Recent advancements demonstrating near or surpassing human intelligence in Transformer-based applications, such as LLMs like GPT-4 (OpenAI, 2024), necessitate a more thorough theoretical understanding. Some researchers are exploring the more general problem of universal computation, specifically the simulation of a universal Turing machine (Schuurmans, 2023). Recently, Schuurmans et al. (2024) presented a more general method for autoregressive generation of universal Turing machine simulations based on an LLM with unaided external memory, albeit assuming arbitrarily long input strings. This contrasts with standard autoregressive generation, where the input is constrained by the context window size. Conversely, Hahn (2020) focuses on the computational limitations of self-attention, a key component of Transformers. Chiang et al. (2022, 2023) tried to overcome the theoretical limitation, and prove a tighter upper bound using a logic. Han's analysis reveals a quadratic time and space complexity relative to sequence length, rendering them inefficient for processing long sequences, particularly in tasks requiring long-range dependency modeling (Merrill et al., 2023).

The mechanisms by which Transformers understand languages are not yet well-established, remaining an active area of research. Strobl et al. (2024) explored the computational power of Transformers from a formal language perspective, investigating "What Formal Languages Can Transformers Express?" Their work focused on expressivity and trainability, proposing a unified framework to reconcile various assumptions that have led to contradictory results. Furthermore, approximation theory (Sanford et al., 2023) explores Transformers as approximators of various classes of functions in connection with the universal approximation theorem.

This article seeks to demystify the enigmatic behaviors of Transformers, offering a practical understanding rooted in formal language theory. Recognizing Transformers as probabilistic approximations of CSLs could significantly advance the field. The seemingly human-like writing and reasoning capabilities of LLMs can be better understood through the lens of classical formal language theory. Notably, many linguistic theorists classify human languages within the CSL category (Chomsky, 1956; Penttonen, 1974; Hopcroft & Ullman,1979).

While extensive research has been conducted on LLM Transformers, varied interpretations often lead to confusion (Choudhary et al., 2022; Singh et al., 2024; chang, 2024; Rai et al.,2024), likely due to differing foundational assumptions. Here, we propose that LLM Transformers can be interpreted as stochastic approximations of left CSLs. Drawing on classical findings, we recall the equivalence between left CSLs and general CSLs, as established by Penttonen (1974): "Every CSL is left context-sensitive". From this perspective, LLMs can be understood as probabilistic implementations of CSLs.

Reframing Transformers as left CSLs within formal language theory provides a rigorous approach to analyzing their error patterns, uncertainty, and sequence modeling capabilities. This article argues that the autoregressive generation behavior of Transformers can be modeled as a stochastic CSL process, emphasizing left context-sensitive grammars (CSGs). Regulated by attention mechanisms, these left CSGs exhibit behaviors akin to regular grammars, offering a cohesive and simplified interpretation of Transformer dynamics. Such a framework helps reconcile apparent contradictions, including the challenges of proving Turing-completeness by Perez et al. (2019) and Hahn's (2020) critical analysis. Moreover, recent studies underscore the limitations of Transformers in capturing specific formal languages, further highlighting the need for this reinterpretation (Strobl et al., 2024; Weiss et al., 2020). This refined perspective holds promise for guiding researchers and engineers in generative AI towards deeper insights and more ambitious applications. The major contributions of this article are:

**Formal Language Interpretation of Next-Token Prediction**
This article frames the next-token prediction mechanism of Transformers as derivation processes within left CSGs. This perspective offers a novel application of formal language theory to analyze LLM generation.

**Reinterpretation of Autoregressive Generation**
We reinterpret the autoregressive generation processes of LLMs through the lens of formal language theories, particularly CSLs. In this framework, attention mechanisms are conceptualized as deep learning-powered search networks that guide the autoregressive generation process.

**Focus on Generative Processes Over Internal Probing**
Unlike recent efforts that examine the internal states of Transformers to understand learned representations and decision-making (Belrose et al., 2023), this article prioritizes interpreting the generative processes themselves. We propose that implicit left CSGs, which are formed through context windows, attention mechanisms, and autoregressive frameworks, govern these processes step by step. This approach aligns with black-box interpretation studies of Transformers, which aim to untangle their complex internal operations.

By focusing on generative processes and separating them from attention mechanisms, this framework offers theoretical advancements and practical applications. Moving beyond the tendency to treat attention and generation as inseparable entities enables deeper insights and more rigorous interpretations.

## 2. RELATED WORKS

Languages can be broadly categorized as natural or formal. Natural languages, like English or Spanish, evolve organically within human societies through usage and repetition. In contrast, formal languages are designed for specific purposes, such as programming computers or expressing logical systems. The natural languages that are known over-context-free, and under-recursive-enumerable. The most reasonable assumption is that they are in context-sensitive class, but there are still many arguments (Wang & Steinert-Threlkeld, 2023).

### 2.1. Formal Language Theory and Natural Language

Formal languages are categorized according to expressive capabilities and defined by the types of production rules, called The Chomsky hierarchy (Chomsky, 1956; Hopcroft & Ullman,1979):

**Type 0 (Recursive Enumerable Language):** Any productions are allowed.
**Type 1 (Context-Sensitive Language):** Production form is $\alpha A \beta \rightarrow \alpha \gamma \beta$.
**Type 2 (Context-Free Language):** Production form is $A \rightarrow \beta$.
**Type 3** (**Regular Language**)**:** Production form is $A \rightarrow a$ or $A \rightarrow aB$.

Natural languages are generally considered to be beyond context-free but below recursively enumerable in complexity. The most widely accepted assumption places them within the context-sensitive class, though ongoing debates and diverse modelling approaches persist. Nakaishi and Hukushima (2024) explored the statistical properties of probabilistic context-sensitive grammars. Nath (2016) examined the relationship between context-sensitive grammars and linear-bounded automata. The existence of supra-context-free in natural language is demonstrated by Swiss German cross-serial dependency (Shieber, 1985a & b) and Bambara vocabulary (Culy, 1985). Weekly equivalent CSG is superior in generative power than natural languages, and efficient parsing is hard problem (Wang & Steinert-Threlkeld, 2023). Tree-Adjoining Grammars (Joshi, 1985) and Combinatory Categorial Grammars (Steedman, 1987) have been proposed to weakness of CSG. Mildly context-sensitive hypothesis is proposed to handle meaningful natural languages restricting the full CSG. The validity of mildly context-sensitive is challenged for more linguistic power (Wang & Steinert-Threlkeld, 2023; Kobele, 2006), but it still remains open question (Clark and Yoshinaka, 2012), and there is no agreement, yet (Graf, 2021). Li et al. (2024) explored the integration of formal language and natural language for developing controllable LLM-based agents. Savitch (1988) investigated the formal complexity of natural language.

### 2.2. Bridging Formal Theory and Transformers

The formal language theory explores Transformers as recognizers or generators of formal structures such as automata, Boolean circuits, or formal logic (Merrill, 2021). Finite state automata are extracted from state space machine learning models for evaluating the automata on formal languages (Weiss, 2020). Zou (2023) simulated the representations of internal Transformers as states of a hypothetical automaton. Automaton extraction can be used for revealing the mechanisms of neural networks, often characterized by their black-box nature (Hong, 2022; Wei, 2024). Zhang et al. (2024) addressed the interpretation of Transformers by extracting deterministic finite automata. Considering Transformers as black boxes (Hassija et al., 2024a), they approximated the state transitions (Weiss, 2020) of Transformers during autoregressive generation processing by deterministic continuous state automata, which are then converted to deterministic finite state automata. Dao & Gu (2024) explored implicit state-like mechanisms (Zhang et al., 2024; Xu et al., 2021) of Transformers. Understanding the dynamics of Transformers in relation to language relies on diverse assumptions, leading to varied results that still require significant research (Strobl et al., 2024).

Studies on Transformers' capability in formal language learning have involved generalization experiments on unseen strings of the same length as the training data. However, the results are often inferior to LSTMs in the class of mildly context-sensitive languages (Wang & Steinert-Threlkeld, 2023). Theoretical explorations of Transformers' capability in formal language recognition present two contradictory views: optimistic and pessimistic interpretations (Sanford et al., 2023). Transformers' ability to recognize counter languages was tested by Bhattamishra et al. (2020b). Yao et al. (2021) investigated the recognition of Dyck languages by Transformers with bounded size and depth. It has been shown that finite state automata with polynomial size using a sequence of length N can be simulated by Transformers of depth log(N) (Liu et al., 2022).

## 2.3. Contrasting LLMs and Formal Languages: Critical and Constructive Perspectives

Chomsky's formal language theory is grounded in the use of axioms and structured hierarchies. It defines formal languages using grammars, where each is governed by explicit production rules. These grammars serve as foundational tools for analyzing the structure of natural languages.

In contrast, LLMs do not rely on predefined formal grammars. Instead, they learn from statistical patterns in vast text corpora. While LLMs can produce grammatically coherent output, their mechanisms differ fundamentally from those of formal language theory. We examine key distinctions between LLMs and formal grammars, offering both pessimistic and optimistic interpretations:

**Data-Driven vs. Rule-Based Approaches**
- **Pessimistic View:** LLMs are fundamentally data-driven, extracting patterns from massive datasets. Formal languages, by contrast, are rule-based systems that use explicitly defined rules to generate and analyze languages.
- **Optimistic View:** Despite their statistical nature, LLMs' next-token predictions, shaped by nonlinear, stochastic functions, can approximate the behavior of formal languages. At specific inference moments, these predictions can be interpreted as context-sensitive production rules, such as those found in left CSLs.

**Flexibility vs. Rigidity**
- **Critical View:** LLMs are praised for their flexibility and creativity in language generation, while formal grammars are often seen as rigid and limited in expressive scope.
- **Constructive View:** Even formal grammar is not inherently statistical, their rules can be dynamically trained and inferred by LLMs. In this way, LLMs can internalize and operate with implicit formal structures, blurring the boundary between fixed rules and adaptive learning.

**Probabilistic vs. Deterministic Models**
- **Critical View:** LLMs operate probabilistically, producing text based on learned likelihoods rather than deterministic rules. This contrasts with formal grammars, which generate output strictly based on fixed production rules. Although LLMs excel at mimicking natural language, their foundations lie in statistical modeling rather than formal linguistic theory.
- **Constructive View:** Formal languages can be generalized to include probabilistic and dynamic variants. In this light, Transformers' next-token predictions can be seen as adaptive production rules, evolving through training and context.

The critical perspectives outlined above stem from viewing LLMs and formal grammars as fundamentally incompatible. However, we argue that these differences can be reconciled through the concept of **stochastic and dynamic production rules**, which are not hardcoded but learned during pretraining. In Section 4, we delve deeper into how these implicit, context-sensitive mechanisms align with formal CSL production rules and can be employed for language derivations.

## 3. PRELIMINAIRES

This section discusses the fundamental principles of basic Transformers, aiming to demystify their essential mechanisms and explore the origins of their apparent intelligence. Departing from conventional approaches that delve into the intricacies of deep neural network layers like attention mechanisms and feed-forward networks, we will instead emphasize essential definitions to isolate the elementary mechanisms at play. This simplified approach aims to provide a clear foundation for understanding the remarkable capabilities of Transformers, while avoiding the complexity that often obscures the basic insights.

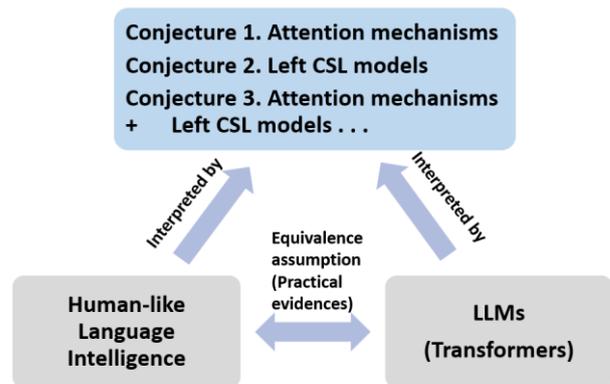

**Figure 1. Possible Conjectures on the Interpretation of LLMs:** While LLMs are widely regarded as exhibiting human-like intelligence in practice, theoretical interpretations of their behavior and mechanisms remain in the early stages of development.

### 3.1. Three Conjectures

This paper explores three potential explanations for the human-like intelligence observed in Transformers (Figure 1):

- **Conjecture 1:** The intelligence stems primarily from the complex deep neural network architecture, particularly the attention

mechanisms, as extensively investigated by numerous researchers (Perez et al., 2019; Hahn, 2020).
- **Conjecture 2:** Transformers essentially perform next-token searches within a context window, dynamically approximating left CSL derivations through autoregressive generation (This article).
- **Conjecture 3:** A combination of Conjectures 1 and 2, where both deep neural network complexity and left CSL approximation contribute to intelligence.

The majority of existing research on Transformer interpretability aligns with Conjecture 1, while very limited studies have explored Conjectures 2 or 3. This article focuses on Conjecture 2. We investigate the human-like intelligence capabilities of standard Transformers by adopting a simplified, 'bare-bones' architectural approach for clear understandings. This simplification allows us to uncover fundamentals, yet often overlooked, insights into the mechanisms underlying Transformer intelligence.

### 3.2. Bare-Bones Transformers

We define three building blocks of bare-bones Transformers: the context window, autoregressive generation architecture, and attention mechanism. Here, the terminology, attention mechanism, is used as the inclusion all layers such as Transformers' self-attentions, forward neural networks, etc. for the simplicity. We begin with basic definitions and terminologies.

**LLMs**: Large Language Models (LLMs) are deep learning models, often built with Transformer architectures, that are pre-trained on substantial datasets. This pre-training enables them to effectively understand and generate human language, with OpenAI's GPT being a prominent example.

**Token:** In language processing, a token is the smallest unit of text, typically a word or a subword, that has been segmented from the input.

**Embedding:** Embedding refers to the process of converting input text and positions of the tokens into a vector representation that captures the semantic meaning of words.

**String:** Concatenation of a finite sequence of tokens, e.g., $\tau_1 \tau_2 \cdots \tau_k$, $k \in \mathbb{N}$, where $\tau_i$ is a token.

**Bare-bones Transformers**: Bare-bones Transformers are deep learning neural networks designed for parallel processing of sequential data (tokens). Utilizing self-attention and positional encodings, they learn context from this data to produce outputs like text and predictions through an autoregressive generation framework. Their fundamental components are attention mechanisms, the context window, and the autoregressive generation process, enabling them to capture dependencies within the context window and generate the next token sequentially.

**Context window:** The context window serves as the working memory of an LLM, storing, accessing, and updating the sequence of tokens (strings). Its maximum length, a practical consideration, has significantly increased in recent models (e.g., GPT-3: 2,049 tokens, GPT-4: 32,768 tokens, GPT-4 Turbo: 128,000 tokens) to effectively capture long-range dependencies.

**Attention Mechanism:** As the central engine of Transformers, the attention mechanism processes the input sequence within the context window to identify dependencies and subsequently predict the next token.

**Autoregressive Generation Process:** The autoregressive generation framework operates iteratively. At each time step, the attention mechanism produces a token, which is subsequently incorporated into the updated context window. This augmented context is then used as the basis for predicting the next token in the sequence. This process continues until a predefined termination condition, such as reaching a target sequence length, is reached.

## 4. TRANSFORMERS ARE STOCHASTIC APPROXIMATIONS OF LEFT CONTEXT-SENSITIVE LANGUAGES

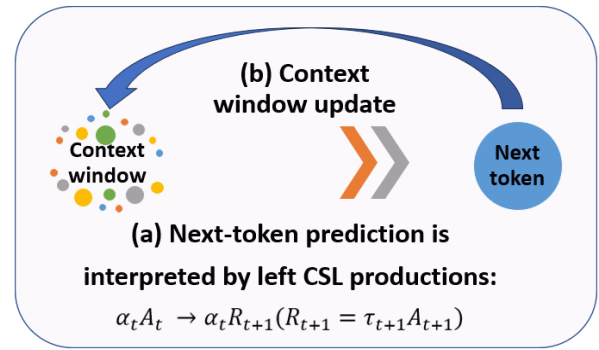

**Figure 2. Interpretation of the next-token prediction as Left CSL production:** (a) Next-token prediction substep, where the context sequence $\alpha_t$ within the context window is fed into the attention mechanism $A_t$ to predict the next token. This predicted token $\tau_{t+1}$ constitutes a variable $R_{t+1}$ in the *right-hand side* of the left CSL production, where $R_{t+1} = \tau_{t+1} A_{t+1}$. Notice that $\alpha_t A_t \rightarrow \alpha_t R_{t+1}$ satisfies the constraint of left CSG (see Eq. (4)). (b) Context window update substep, where the context sequence is then updated to include the predicted token. By repeatedly applying substeps (a) and (b), the Transformer's autoregressive generation process can be interpreted as a sequence of left CSL derivation (see Eq. (5)).

Much research has focused on probing the internal states of Transformers to understand their learned representations and decision-making processes (Zhao et al., 2024a & b; Bartoszcze et al., 2025). Rai et al. (2024)

presents a comprehensive survey of mechanistic interpretability approaches for Transformer-based LLMs, focusing on reverse-engineering internal model computations. The authors adopt a task-centric lens, highlighting both novel insights into model behavior and the practical challenges that arise when interpreting complex deep learning systems. Even through a lot of research efforts have investigated on probing and interpretability of internal mechanisms of Transformers (Bereska & Gavves, 2024; Dong, 2023) a significant challenge in understanding LLM Transformers lies in the conflation of attention mechanisms and autoregressive generation.

We propose a separation of these processes, viewing attention mechanisms as deep learning-based search networks that guide the autoregressive generation (Fig. 2). This generation process, in turn, is governed by implicit left CSGs decided by the context window and attention mechanism at each prediction step. This separation aligns with black-box interpretation studies of Transformers, which seek to disentangle their complex internal operations (Mahinpei et al., 2021; Choudhary et al., 2022. Hassija et al., 2024b; Şahin et al., 2024; Sam & Finzi, 2024; Sam et al., 2025).

### 4.1. Left CSL and Autoregressive Generation

Our central hypothesis is that Transformers can be decomposed into three distinct building blocks: the context window, attention mechanism, and autoregressive generation process frameworks. This separation facilitates the exploration of more flexible computational models and optimization strategies for Transformer autoregressive processes, moving beyond the limitations of treating attentions and autoregressive generations as inseparable entities. We interpret the next-token predictions of Transformers as the probabilistic approximations of left CSL productions, and the autoregressive generation as the derivation of left CSLs, providing a formal language theory perspective.

After tokenization, an autoregressive generation process can be described by the updates of the context windows and the state transitions of Transformer attention mechanisms. Let's define the configuration of a Transformer at time $t$ as:

$$(\alpha_t, AM_t),$$

where $\alpha_t$ and $AM_t$ denote the states of the context window and attention mechanism at time step $t$, respectively. The intermediate configuration of a Transformer, occurring after a next-token prediction and before the context window is updated, is represented as:

$$(\alpha_t, \tau_{t+1}, AM_{t+1}),$$

where $\tau_{t+1}$ indicates a predicted next token. Binary relation $\vdash$ is defined as one step or intermediate substep transition between Transformer configuration. The relation $\vdash^*$ means that the transitions of Transformer configuration are zero or more steps and is the reflexive transitive closure of $\vdash$. Then, autoregressive generation process can be represented by:

$$(\alpha_0, AM_0) \vdash (\alpha_0, \tau_1, AM_1) \vdash (\alpha_1, AM_1) \cdots (\alpha_t, AM_t)$$
$$\vdash (\alpha_t, \tau_{t+1}, AM_{t+1}) \vdash \alpha_{t+1}, AM_{t+1}) \cdots \vdash (\alpha_T, AM_T). \quad (1)$$

At time step $t$, attention mechanism $AM_t$ inputs the sequence in context window $\alpha_t$, and performs next-token prediction, and outputs next token $\tau_{t+1}$. The state of the attention mechanism is transited to $AM_{t+1}$. Notice that the context window is still $\alpha_t$ before updated in the intermediate substep. In the next intermediate substep, the context window is updated to $\alpha_{t+1}$. $\alpha_0$ denotes the initial context window which is an initial sequence, i.e., a given prompt and blank symbols, if necessary. $\alpha_T$ shows a final generated sequence.

For clarity, we distinguish the configuration transitions into the next-token prediction and context window update sub-steps: $\vdash^{ntp}$ indicates the next-token prediction sub-step, and $\vdash^{cwu}$ indicates the context window update sub-step. Eq. (1) can be rewritten as:

$$(\alpha_0, AM_0) \vdash^{ntp} (\alpha_0, \tau_1, AM_1) \vdash^{cwu} (\alpha_1, AM_1) \cdots$$
$$(\alpha_t, AM_t) \vdash^{ntp} (\alpha_t, \tau_{t+1}, AM_{t+1}) \vdash^{cwu} (\alpha_{t+1}, AM_{t+1})$$
$$\cdots \vdash^{cwu} (\alpha_T, AM_T). \quad (2)$$

The preceding observations can be interpreted as a left CSL derivation process, as elaborated in the following section. To further enhance our understanding of Transformer behavior, we establish a bridge between Transformers and formal language theory.

### 4.2. Transformers are Dynamic and Stochastic Approximations of Left CSLs

We argue that next-token predictions can be explained as the probabilistic approximations of left CSL productions, which give intuition how the simple one-token predictions can achieve huma-like intelligence. We will follow the notations of Salomaa (1973) for formal language representation. Let a **variable** $V = V_N \cup V_T$, where $V_N$ and $V_T$ are **nonterminal** and **terminal** symbols (tokens here), respectively, and $V_N$ and $V_T$ are disjoint sets. A production is context-sensitive iff it is of the form:

$$\alpha A \beta \rightarrow \alpha R \beta, \quad (3)$$

where $\alpha, \beta \in V^*$, $R \in V^+$, and $A \in V_N$. A CSG is an ordered quadruple:

$$H = (V_N, V_T, B, Q),$$

where $B \in V_N$ is the start symbol, and $Q$ is a set of context-sensitive productions. A production is left context-sensitive iff it is of the form:

$$\alpha A \to \alpha R, \quad (4)$$

where $\alpha \in V^*$, $R \in V^+$, and $A \in V_N$. A left CSL grammar is an ordered quadruple:

$$G = (V_N, V_T, B, P),$$

where $B \in V_N$ is the start symbol, and $P$ is a set of left context-sensitive productions. We define the language generated by a grammar $G$:

$$L(G) = \{w | B \Rightarrow^* w\} \cap V_T^*,$$

where $w \in V^*$. The relation $\Rightarrow^*$ means that the derivation of $G$ is zero or more steps and is the reflexive transitive closure of $\Rightarrow$. Binary relation $\Rightarrow$ means one step derivation of $G$ on string in $V^*$ is defined by:

$$w \Rightarrow y \text{ iff } \exists u, v, p, q \in V^*:$$
$$(w = upv) \wedge (p \to q \in P) \wedge (y = uqv)$$

Although formal languages do not possess predefined variables in the conventional sense, LLMs can be regarded as providers of *implicit variables* within language derivations. These implicit variables are analogous to the configuration states present in LLMs during generation.

**Hypothesis 1:** Given the configurations of Transformer models during autoregressive generation, the dynamic states of these configurations are assumed to be systematically associated with the dynamic variables of formal language grammar.

We introduce dynamic **nonterminals**, denoted $A_t$ ($t = 0, \ldots, T$-1) corresponds to the attention mechanism states, $AM_t$ as defined in Eqs. (1) and (2). To bridge the gap between formal language theory and the autoregressive generation process of Transformers, we interpret $\alpha_t$ as a string (variable). This string can be viewed as the concatenation of tokens present in the context window at time step $t$. In particular, $\alpha_0$ denotes the initial string, consisting of either a sequence of tokens or blank placeholders within the Transformer's context window. The final string $\alpha_T$, is the complete output generated within the context window, or denoted by $w \in V^*$ as a language generated by the left CSG. Considering Eq. (1), the derivation (or generation) of $w$ can be expressed as follows:

$$B \Rightarrow \alpha_0 A_0 \Rightarrow^* \alpha_1 A_1 \cdots \alpha_t A_t \Rightarrow^* \alpha_{t+1} A_{t+1} \cdots$$
$$\Rightarrow^* \alpha_T A_T \Rightarrow \alpha_T (= w) \quad (5)$$

For clarity, we decompose the derivation process into two distinct steps: next-token prediction, denoted by $\Rightarrow^{ntp}$ and context window update, denoted by $\Rightarrow^{cwu}$. To formalize the derivation steps involving context window updates, we focus on the production rules governing the derivation. Specifically, we examine the detailed derivation steps from the initial variable and nonterminal, denoted by $\alpha_0 A_0$ to the configuration, $\alpha_1 A_1$ (as expressed by $\alpha_0 A_0 \Rightarrow^* \alpha_1 A_1$):

$$\alpha_0 A_0 \Rightarrow^{ntp} \alpha_0 R_1 \Rightarrow^{cwu} \alpha_1 A_1,$$

where $R_1 = \tau_1 A_1$. In general, at time $t$ the detail derivation steps of $\alpha_t A_t \Rightarrow^* \alpha_{t+1} A_{t+1}$ is:

$$\alpha_t A_t \Rightarrow^{ntp} \alpha_t R_{t+1} \Rightarrow^{cwu} \alpha_{t+1} A_{t+1},$$

where $R_{t+1} = \tau_{t+1} A_{t+1}$ ($t = 0, \ldots, T$-1).

We define production rules used for each substep in the next-token prediction process dynamically, including both the initial and final steps of the derivation.

$$\begin{array}{c} B \to \alpha_0 A_0 \\ \alpha_0 A_0 \to \alpha_0 R_1 (R_1 = \tau_1 A_1) \\ \cdots \\ \alpha_t A_t \to \alpha_t R_{t+1} (R_{t+1} = \tau_{t+1} A_{t+1}) \quad (6) \\ \cdots \\ \alpha_{T-1} A_{T-1} \to \alpha_{T-1} R_T (R_T = \tau_T A_T) \\ A_T \to \lambda \end{array}$$

Notice that productions $\alpha_0 A_0 \to \alpha_0 R_1$, $\cdots$, $\alpha_{T-1} A_{T-1} \to \alpha_{T-1} R_T$ in Eq. (6) satisfy the constraints of the left CSL production (see Eq. (4)).

Thereafter, the context window update works as follows:
1. The procedure starts with a variable $R_{t+1}$, which is split into two parts: $\tau_{t+1}$ and $A_{t+1}$.
2. It then takes the current context $\alpha_t$ along with $\tau_{t+1}$ as inputs to produce the updated context $\alpha_{t+1}$, which is saved in the context window.
3. Essentially, the new configuration $\alpha_{t+1} A_{t+1}$ is obtained from the previous configuration $\alpha_t R_{t+1}$.
4. Since these transitions do not follow strict formal language production rules, we call them pseudo derivations, denoted by:

$$\alpha_t R_{t+1} \Rightarrow^{cwu} \alpha_{t+1} A_{t+1}.$$

The pseudo derivation steps don't produce language (i.e., tokens) themselves, rather, they are responsible for transitioning the variables used in language derivations. Since this idea is intuitively clear, we omit the formal proof as it lies beyond the scope of this article. Consequently, Eq. (5) is not a pure language derivation step in the sense of formal language theory. However, we argue that the next-token prediction in Eq. (5) can be seen as a probabilistic, dynamic approximation of left context-sensitive language (CSL) productions. This perspective offers an intuitive explanation for how such predictions can exhibit human-like language intelligence. Building on this idea, we argue that

Transformers function as dynamic, stochastic approximations of left CSLs, since the attention mechanisms underlying next-token prediction can be viewed as stochastic approximators. In essence, Transformers are trained through next-token prediction and implicitly emulate the generative behaviour of left CSLs.

**Theorem 1:** Autoregressive generations in Transformers can be interpreted as left CSL derivations.

**PF>** If we conceptualize the role of attention mechanisms as conducting next-token searches within the current context window, then the autoregressive generation process of Transformer-based models can be regarded as a stochastic and dynamic approximation of left CSL derivations. The analysis presented in Section 4.2 substantiates this perspective, demonstrating that the generative processes of LLMs align structurally with derivations characteristic of left CSLs.

### 4.3. Transformers are CSL Generators

It is intuitively evident that left CSLs exhibit less expressive power than general context-sensitive languages, whereas a theoretical framework widely regarded as a plausible model for human language. n this context, the production rules governing left CSLs represent a natural restriction of those for general CSLs (Havel, 1969; Havel, 1970). Katz (1974) conducted a seminal study on synchronized left context-sensitive transformations of regular languages. Whether the restriction reduces language generating powers, had been solved by the study of Penttonen (1974).

**Theorem 2** (Penttonen, 1974)**:** Every CSL is left context-sensitive.

**PF>** It is proven that we can simulate any context-sensitive grammar by left context-sensitive productions (Penttonen, 1974).

Since Penttonen (1974) has established the equivalence between context-sensitive languages (CSLs) and left context-sensitive languages (left CSLs), we conclude that Transformers function as stochastic approximations of CSLs. Consequently, Transformers are naturally accepted as human-like intelligence generators.

**Theorem 3:** Transformers are CSL generators.

**PF>** This result follows directly from Theorems 1 and 2.

### 4.4 Computation Efficiency and Future Research Direction

LLMs have been widely recognized as exhibiting human-like intelligence since the emergence of GPT-3 (Brown, 2020; Goldstein et al., 2025). Theorem 3 offers a deeper understanding and valuable insight into how Transformers have achieved such remarkable generative AI performance, surpassing, in some cases, human-level intelligence (Chang et al., 2023). While some researchers analyzing the layered neural structures of attention mechanisms have reported pessimistic conclusions (Han, 2020), a growing body of experimental evidence has consistently demonstrated that LLMs can produce human-like, or even superior, intelligence since the release of GPT-3 (OpenAI, 2023; Brown, 2020).

Our novel perspective on Transformer architecture seeks to foster a deeper understanding of LLMs and their future potential in terms of both qualitative and quantitative throughput. Left CSGs inherently produce skewed derivation trees, which can be computationally inefficient. We propose that exploring expansions to general forms of CSGs within autoregressive generation could yield substantial improvements in both performance and computational resource utilization.

## 5. CONCLUDING REMARKS

We argue that next-token predictions can be interpreted as probabilistic, dynamic approximations of left CSL productions. This perspective provides an intuitive explanation for how seemingly simple token predictions can yield outputs that exhibit human-like intelligence. Given that all CSLs are left context-sensitive (Penttonen, 1974), we conclude that Transformers stochastically approximate CSLs, which are widely recognized as formal models of aspects of human-like intelligence.

This work also aims to bridge the gap between formal language theory and the practical generative power of Transformers, laying a foundation for future advancements in generative AI theory and applications. Traditionally, improvements in LLM capabilities have been primarily attributed to increases in the size of network parameters and the extent of training data. Indeed, scaling has significantly enhanced LLM performance. However, the sheer size and complexity of state-of-the-art models make them difficult to verifiably control and modify (Bhattamishra et al., 2021). Recent work suggests that advancement does not *strictly* depend on scaling, but also arises from architectural innovations, training techniques, and the ability to generalize across tasks (DeepSeek-AI et al., 2024).

## REFERENCES

Bartoszcze, L., Munshi, S., Sukidi, B., Yen, J., Yang, Z., Williams-King, D., Le, L., Asuzu, K., & Maple, C. (2025). Representation Engineering for Large-Language Models: Survey and Research Challenges. arXiv. Retrieved from http://arxiv.org/abs/2502.17601.

Belrose, N., Furman, Z., Smith, L., Halawi, D., Ostrovsky, I., McKinney, L., Biderman, S., Steinhardt, J. (2023). Eliciting Latent Predictions from Transformers with the Tuned Lens. arXiv:2303.08112v4 [cs.LG] 26 Nov 2023. http://arxiv.org/abs/2303.08112.

Bender, E. M., Gebru, T., McMillan-Major, A., & Shmitchell, S. (2021). On the dangers of stochastic parrots: Can language models be too big? In Proceedings of the 2021 ACM conference on fairness, accountability, and transparency (pp. 610–623).

Bereska, L., & Gavves, E. (2024). Mechanistic Interpretability for AI Safety–A Review. arXiv preprint arXiv:2404.14082.

Bhattamishra, S., et al. (2020a). On the computational power of transformers and its implications in sequence modeling. In Conference on Computational Natural Language Learning (CONLL). https://doi.org/10.18653/v1/2020.conll-1.37 Retrieved from https://www.aclweb.org/anthology/2020.conll-1.37

Bhattamishra, S., Ahuja, K., & Goyal, N. (2020b). On the Ability and Limitations of Transformers to Recognize Formal Languages. arXiv:2009.11264 [cs.CL].

Brown, T., Mann, B., Ryder, N., Subbiah, M., Kaplan, J. D., Dhariwal, P., Neelakantan, A., Shyam, P., Sastry, G., Askell, A., et al. (2020). Language models are few-shot learners. Advances in Neural Information Processing Systems, 33, 1877–1901.

Chang, Y. et al. (2023). A Survey on Evaluation of Large Language Models. arXiv:2307.03109v9 [cs.CL] 29 Dec 2023.

Chiang, D., & Cholak, P. (2022). Overcoming a theoretical limitation of self-attention. In Proceedings of the 60th Annual Meeting of the Association for Computational Linguistics (Volume 1: Long Papers) (pp. 7654–7664). Association for Computational Linguistics.

Chiang, D., Cholak, P., & Pillay, A. (2023). Tighter bounds on the expressivity of transformer encoders. In Proceedings of the 40th International Conference on Machine Learning, 202 of Proceedings of Machine Learning Research (pp. 5544–5562). PMLR.

Chomsky, N. (1956). Three models for the description of language. IRE Transactions on Information Theory, 2(3), 113-124.

Choudhary, S., Chatterjee, N., & Saha, S. K. (2022). Interpretation of black box nlp models: A survey. arXiv preprint arXiv:2203.17081.

Clark, A., & Yoshinaka, R. (2012). Beyond semilinearity: Distributional learning of parallel multiple context-free grammars. In International Conference on Grammatical Inference (pp. 84–96). PMLR.

Culy, C. (1985). The Complexity of the Vocabulary of Bambara. In The Complexity of Lexical Entries and Rules (pp. 349–357). Springer Netherlands.

Dao, T., & Gu, A. (2024). Transformers are SSMs: Generalized Models and Efficient Algorithms Through Structured State Space Duality. arXiv:2405.21060.

DeepSeek-AI, Liu, A., et al. (2024). DeepSeek-V3 Technical Report. arXiv:2412.19437 [cs.CL]. Retrieved from https://arxiv.org/abs/2412.19437.

Devlin, J., et al. (2019). BERT: Pre-training of Deep Bidirectional Transformers for Language Understanding. arXiv:1810.04805 [cs.CL].

Dong, X., Wang, Y., Yu, P. S., & Caverlee, J. (2023). Probing explicit and implicit gender bias through llm conditional text generation. arXiv preprint arXiv:2311.00306.

Goldstein, S., & Levinstein, B. A. (2025). Does ChatGPT Have a Mind? arXiv preprint arXiv:2407.11015.

Google. (2025). "Gemini 2.5: Our most intelligent AI model". Google DeepMind. Retrieved April 8, 2025.

Graf, T. (2021). Minimalism and computational linguistics. Lingbuzz/005855. Handbook of Minimalism.

Grzankowski, A., et al. LLMs are Not Just Next Token Predictors. arXiv:2408.04666 [cs.CL]. https://doi.org/10.48550/arXiv.2408.04666.

Hahn, M. (2020). Theoretical Limitations of Self-Attention in Neural Sequence Models. Transactions of the Association for Computational Linguistics, 8, 156–171.

Hassija, V., et al. (2024a). Interpreting Black-Box Models: A Review on Explainable Artificial Intelligence. Cognitive Computation, 16, 45–74.

Hassija, V., et al. (2024b). Survey- the development of XAI is reviewed meticulously through careful selection and analysis of the current state-of-the-art of XAI research.

Havel, I. (1969). A Note on One-Sided Context-Sensitive Grammars. Kybernetika, Vol. 5 (1969), No. 3, (186)—189. URL: http://dml.cz/dmlcz/125848.

Havel, I. (1970). On one-sided context-sensitive grammars. In J. Dörr & G. Hotz (Eds.), Automatentheorie und Formale Sprachen. Bibliographisches Institut. Moscow (1970), pp. 221-225.

Hong, D., Segre, A. M., & Wang, T. (2022). Adaax: Explaining recurrent neural networks by learning automata with adaptive states. In Proceedings of the 28th ACM SIGKDD Conference on Knowledge Discovery and Data Mining (pp. 574–584).

Hopcroft, J. E., & Ullman, J. D. (1979). Introduction to automata theory, languages, and computation. Addison-Wesley.

Joshi, A. K. (1985). Tree-adjoining grammars: How much context-sensitivity is required to provide reasonable structural descriptions? Studies in Natural Language Processing, 206–250. Cambridge University Press.

Katz, B. E. (1974). Synchronized left context-sensitive transformations of regular languages. Nauch. Tehh. Inform. Ser. 2, 1974(4), 36.

Kobele, G. M. (2006). Generating copies: An investigation into structural identity in language and grammar (Doctoral dissertation, University of California, Los Angeles).

Li, Y., Zhao, W., Shen, C., & Chen, J. (2023). A Survey on Evaluation of Large Language Models. arXiv preprint arXiv:2307.03109.

Li, Z., et al. (2024). Formal-LLM: Integrating Formal Language and Natural Language for Controllable LLM-based Agents. arXiv:2402.00798v2 [cs.LG].

Liu, B., Ash, J., Goel S., Krishnamurthy A., Zhang C. (2022). Transformers Learn Shortcuts to Automata. arXiv:2210.10749v2 [cs.LG] 2 May 2023.

Liu, Y., Zhong, M., Xu, R., Zhu, J., Zhang, Y., et al. (2023). Recent Advances in Large Language Models: A Survey. arXiv preprint arXiv:2307.06435.

Mahinpei, A., Clark, J., Lage, I., Doshi-Velez, F., & Pan, W. (2021). Promises and pitfalls of black-box concept learning models. arXiv preprint arXiv:2106.13314.

Merrill, W. (2021). Formal Language Theory Meets Modern NLP. http://arxiv.org/abs/2102.10094.

Merrill, W., Sabharwal, A., & Clark, C. (2023). Modeling Long-Range Dependencies: A Survey of Recent Advances. arXiv preprint arXiv:2308.10623.

Meta. (2025). The Llama 4 herd: The beginning of a new era of natively multimodal AI innovation. ai.meta.com. Archived from the original on 2025-04-05. Retrieved 2025-04-05.

Nakaishi, K., & Hukushima, K. (2024). Statistical properties of probabilistic context-sensitive grammars. arXiv:2402.07113v3 [cond-mat.dis-nn].

Nath, P. (2016). Context-Sensitive Grammars and Linear-Bounded Automata. International Journal of Computer Network and Information Security. I. J. Computer Network and Information Security, 2016, 1, 61-66. DOI: 10.5815/ijcnis.2016.01.08.

OpenAI. 2023. https://chat.openai.com.chat.

OpenAI. (2024). GPT-4 v6 Technical Report. arXiv preprint arXiv:2303.08774.

Penttonen, M. (1974). One-Sided and Two-Sided Context in Formal Grammars. Information and Control, 25, 371–392.

P´erez, J., Marinkovi´c, J., & Barcel´o, P. (2019). On the Turing completeness of modern neural network architectures. In International Conference on Learning Representations (ICLR).

Rai, D., Zhou, Y., Feng, S., Saparov, A., & Yao, Z. (2024). A practical review of mechanistic interpretability for transformer-based language models. arXiv preprint arXiv:2407.02646.

Şahin, E., Arslan, N. N., & Özdemir, D. (2024). Unlocking the black box: an in-depth review on interpretability, explainability, and reliability in deep learning. Neural Computing and Applications, 1–107.

Salomaa, A. (1973). Formal Languages. Academic Press.

Sam, D., & Finzi, M. A. (2024). Eliciting Black-Box Representations from LLMs through Self-Queries. In ICML 2024 Next Generation of AI Safety Workshop.

Sam, D., Finzi, M., & Kolter, J. Z. (2025). Predicting the Performance of Black-box LLMs through Self-Queries. arXiv preprint arXiv:2501.01558.

Sanford, C., et al. (2023). Representational Strengths and Limitations of Transformers. arXiv:2306.02896v2 [cs.LG] 16 Nov 2023.

Sanghai, N., & Brown, N. B. (2024). Advances in Transformers for Robotic Applications: A Review. arXiv:2412.10599v1 [cs.RO].

Savitch, W. J. (1988). THE FORMAL COMPLEXITY OF NATURAL LANGUAGE. Computational Linguistics, 14(4), 231-242.

Schuurmans, D. (2023). Memory augmented large language models are computationally universal. arXiv 2301.04589.

Schuurmans, D., et al. (2024). Autoregressive Large Language Models are Computationally Universal. arXiv:2410.03170v1 [cs.CL].

Shanahan, M. (2024). Talking about large language models. Communications of the ACM, 67(2), 68-79.

Shieber, S. M. (1985a). Evidence Against the Context-Freeness of Natural Language. In Readings in Natural Language Processing (pp. 320–334). Morgan Kaufmann Publishers Inc.

Shieber, S. M. (1985b). Evidence against the context-freeness of natural language. In Philosophy, Language, and Artificial Intelligence (pp. 79–89). Springer.

Singh C., Inala J., Galley M., Caruana R., Gao J. 2024. Rethinking Interpretability in the Era of Large Language Models. arXiv:2402.01761v1 [cs.CL] 30 Jan 2024.

Steedman, M. (1987). Combinatory grammars and parasitic gaps. Natural Language and Linguistic Theory, 5(3), 403–439.

Strobl, L., Merrill, W., Weiss, G., Chiang, D., & Angluin, D. (2024). What Formal Languages Can Transformers Express? A Survey. arXiv:2311.00208v3 [cs.LG].

Vaswani, A., Shazeer, N., Parmar, N., Uszkoreit, J., Jones, L., Gomez, A. N., Kaiser, Ł., Polosukhin, I. (2017). Attention is all you need. Advances in Neural Information Processing Systems, 30.

Wang, S., & Steinert-Threlkeld, S. (2023). Evaluating Transformer's Ability to Learn Mildly Context-Sensitive Languages. In Proceedings of the 6th BlackboxNLP Workshop: Analyzing and Interpreting Neural Networks for NLP (pp. 271–283).

Wei, Z., Zhang, X., Zhang, Y., & Sun, M. (2024). Weighted automata extraction and explanation of recurrent neural networks for natural language tasks. Journal of Logical and Algebraic Methods in Programming, 136, 100907.

Weiss, G., Goldberg, Y., & Yahav, E. (2020). Extracting Automata from Recurrent Neural Networks Using Queries and Counterexamples. arXiv:1711.09576 [cs.LG].

Xu, Z., Wen, C., Qin, S., & He, M. (2021). Extracting automata from neural networks using active learning. PeerJ Computer Science, 7, e436.

Yao, S., Peng, B., Papadim-itriou, C., & Narasimhan, K. (2021). Self-attention networks can process bounded hierarchical languages. In Proceedings of the 59th Annual Meeting of the Association for Computational Linguistics and the 11th International Joint Conference on Natural Language Processing (ACL-IJCNLP) (pp. 3770–3785).

Shukang Yin, Chaoyou Fu, Sirui Zhao, Ke Li, Xing Sun, Tong Xu, and Enhong Chen. A Survey on Multimodal Large Language Models. arXiv:2306.13549v4 [cs.CV] 29 Nov 2024.

Zhang, Y., Wei, Z., & Sun, M. (2024). Automata Extraction from Transformers. arXiv:2406.05564v1 [cs.LG].

Zhao, H., Chen, H., Yang, F., Liu, N., Deng, H., Cai, H., Wang, S., Yin, D., & Du, M. (2024a). Explainability for Large Language Models: A Survey. ACM Transactions on Intelligent Systems and Technology, 15(2), Article 20. https://doi.org/10.1145/3639372

Zhao, H., Yang, F., Lakkaraju, H., & Du, M. (2024b). Opening the black box of large language models: Two views on holistic interpretability. arXiv e-prints, arXiv–2402.

Zou, A., Phan, L., Chen, S., Campbell, J., Guo, P., Ren, R., Pan, A., Yin, X., Mazeika, M., Dombrowski, A.-K., Goel, S., Li, N., Byun, M. J., Wang, Z., Mallen, A., Basart, S., Koyejo, S., Song, D., Fredrikson, M., Kolter, J. Z., & Hendrycks, (2023). Representation Engineering: A Top-Down Approach to AI Transparency. arXiv:2310.01405 [cs.LG]